\pdfoutput=1

\documentclass[11pt]{article}

\usepackage{acl}

\usepackage{times}
\usepackage{latexsym}

\usepackage[T1]{fontenc}

\usepackage[utf8]{inputenc}

\usepackage{microtype}

\usepackage{inconsolata}

\usepackage{graphicx}

\usepackage{xspace}
\usepackage{inconsolata}
\usepackage{booktabs}
\usepackage{comment}
\usepackage{tcolorbox} 
\usepackage{framed}
\usepackage{CJKutf8} 
\usepackage{multirow}
\usepackage{multicol}
\usepackage{amsmath}
\usepackage{amssymb}
\usepackage{amsfonts}
\DeclareMathOperator*{\argmax}{arg\,max}

\def\figref#1{Figure ~\ref{fig:#1}}
\def\figlabel#1{\label{fig:#1}\label{p:#1}}

\def\tabref#1{Table ~\ref{tab:#1}}

\def\tablabel#1{\label{tab:#1}\label{p:#1}}

\def\secref#1{\S\ref{sec:#1}}
\def\seclabel#1{\label{sec:#1}}

\def\appref#1{Appendix ~\ref{app:#1}}
\def\applabel#1{\label{app:#1}\label{p:#1}}

\newcommand\blfootnote[1]{%
  \begingroup
  \renewcommand\thefootnote{}\footnote{#1}%
  \addtocounter{footnote}{-1}%
  \endgroup
}

\title{BMIKE-53: Investigating Cross-Lingual Knowledge Editing with In-Context Learning}

\author{Ercong Nie$^\ast$\textsuperscript{1,2}\qquad Bo Shao$^\ast$\textsuperscript{3,6} \qquad Mingyang Wang\textsuperscript{1,2,4}  \\ 
~\textbf{Zifeng Ding\textsuperscript{1,2,5}\qquad Helmut Schmid\textsuperscript{1}}\qquad \textbf{~Hinrich Sch\"utze\textsuperscript{1,2}}\\
\textsuperscript{1}LMU Munich, Germany \qquad \textsuperscript{2}Munich Center for Machine Learning, Germany \\
\textsuperscript{3}Technical University of Munich, Germany \qquad
\textsuperscript{4}Bosch Center for Artificial Intelligence, Germany
\\
\textsuperscript{5}University of Oxford, UK \qquad
 \textsuperscript{6}CISPA Helmholtz Center for Information Security, Germany\\
\texttt{nie@cis.lmu.de} \qquad \texttt{bo.shao@tum.de}\\
}

\begin{document}
\maketitle



\begin{abstract}
  \blfootnote{$^\ast$~Equal contributions.}

This paper introduces BMIKE-53, a comprehensive benchmark for cross-lingual in-context knowledge editing (IKE) across 53 languages, unifying three knowledge editing (KE) datasets: zsRE, CounterFact, and WikiFactDiff. Cross-lingual KE, which requires knowledge edited in one language to generalize across others while preserving unrelated knowledge, remains underexplored. To address this gap, we systematically evaluate IKE under zero-shot, one-shot, and few-shot setups, incorporating tailored metric-specific demonstrations. Our findings reveal that model scale and demonstration alignment critically govern cross-lingual IKE efficacy, with larger models and tailored demonstrations significantly improving performance. Linguistic properties, particularly script type, strongly influence performance variation across languages, with non-Latin languages underperforming due to issues like language confusion.
Code and data are publicly available at: \url{https://github.com/ercong21/MultiKnow/}.

\end{abstract}

\section{Introduction}
Large language models (LLMs) have demonstrated remarkable abilities to encode vast amounts of knowledge during pre-training, enabling them to perform well across a range of tasks~\citep{min-etal-2022-rethinking,zhang2023r,zhou2024lima}. However, this knowledge remains static, becoming outdated as the world evolves, necessitating mechanisms to update models with new facts while preserving their overall performance~\citep{cao-etal-2021-knowledgeable,dhingra-etal-2022-time}. Traditional approaches, such as fine-tuning, are computationally expensive and impractical for closed-source or large-scale models~\citep{dai-etal-2022-knowledge}. These limitations have motivated the emergence of knowledge editing (KE)---a technique for selectively modifying LLMs to incorporate new knowledge while maintaining the integrity of unrelated knowledge~\citep{zhang-etal-2024-knowledge}.

Recent advancements in KE have explored gradient-free methods inspired by in-context learning (ICL), where LLMs learn through prompts and demonstrations without requiring parameter updates~\citep{zheng2023can}. These methods are efficient and particularly suitable for scenarios where direct access to model parameters is restricted.
However, existing gradient-free KE research primarily focuses on monolingual settings, leaving the potential for cross-lingual KE largely unexplored. Cross-lingual KE, a more challenging task, as illustrated in \figref{figure1}, requires knowledge edited in one language (e.g., English) to generalize effectively to semantically equivalent queries across diverse target languages while preserving unrelated knowledge. 

\begin{figure}[t]
    \centering
    \includegraphics[width=\linewidth]{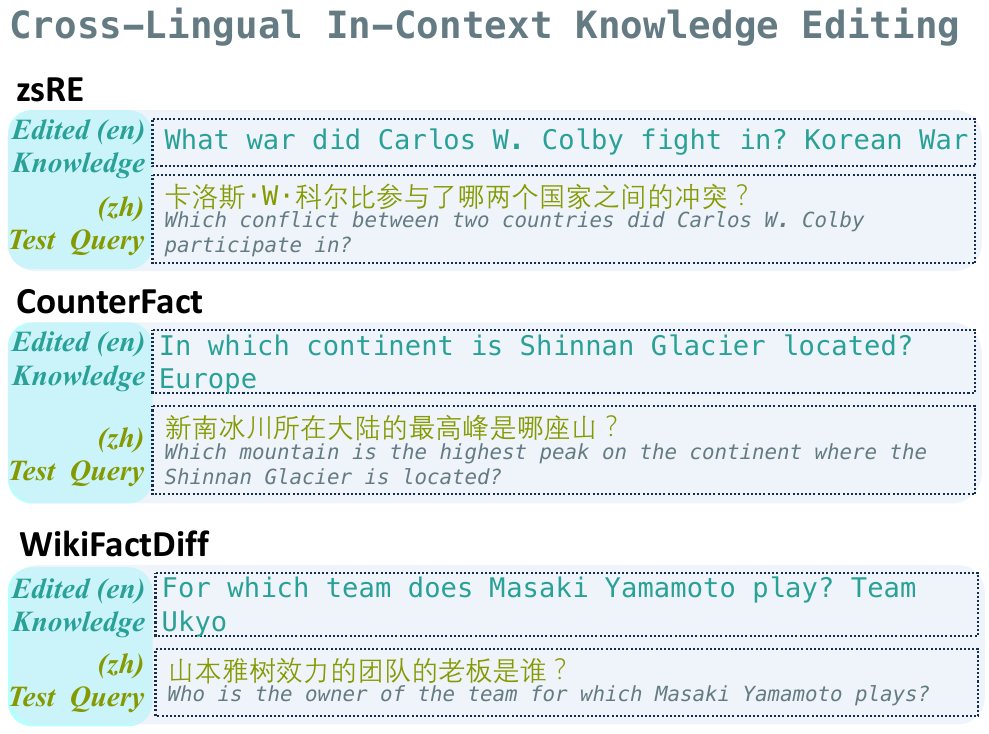}
    \caption{Examples of cross-lingual in-context knowledge editing.}
    \figlabel{figure1}
\end{figure}

This study addresses the critical gap in cross-lingual knowledge editing by proposing a comprehensive and multidimensional investigation of in-context knowledge editing (IKE) methods. We introduce \textbf{BMIKE-53}, a multilingual benchmark spanning 53 languages and integrating three representative KE datasets: zsRE, which evaluates regular fact modifications; CounterFact, which examines counterfactual knowledge updates; and WikiFactDiff (WFD), which assesses real-world, temporally dynamic knowledge updates. This benchmark is the most comprehensive multilingual KE resource to date, unifying diverse KE datasets into a consistent format and expanding them into multiple languages using LLM-assisted translation. The wide linguistic coverage allows us to analyze cross-lingual differences and their underlying causes systematically.

To evaluate cross-lingual IKE, we implement zero-shot, one-shot, and few-shot setups to explore the impact of demonstration quality and quantity on performance~\figref{ike_pipeline}. Notably, we propose two few-shot setups: 8-shot mixed demonstrations, which expose the model to diverse query types, and 8-shot metric-specific demonstrations, which target specific query types like locality or portability to enhance performance. These setups allow us to analyze the interplay between demonstration strategies, query types, and cross-lingual transfer. Our findings show that larger models and tailored demonstrations significantly improve performance, especially for complex queries. Linguistic properties, such as syntactic and phonological similarity with English, positively influence performance, while language family has no significant impact. Instead, script type emerges as a critical factor, with non-Latin languages underperforming due to issues like language confusion, where models generate answers in English instead of the target language.

In summary, our contributions are as follows: \\
\indent \textbf{i)} We introduce BMIKE-53, the most comprehensive multilingual KE benchmark, covering 53 languages and three diverse KE datasets, which serves as a foundation for evaluating cross-lingual KE methods. \\
\indent \textbf{ii)} We extensively evaluate gradient-free cross-lingual KE methods under various IKE setups, providing valuable insights into the effectiveness of in-context learning for cross-lingual knowledge editing. \\
\indent \textbf{iii)} We conduct a detailed analysis of factors influencing cross-lingual KE performance, uncovering the impact of linguistic properties, script types, and language confusion on cross-lingual knowledge transfer.

\begin{figure*}
    \centering
    \includegraphics[width=\linewidth]{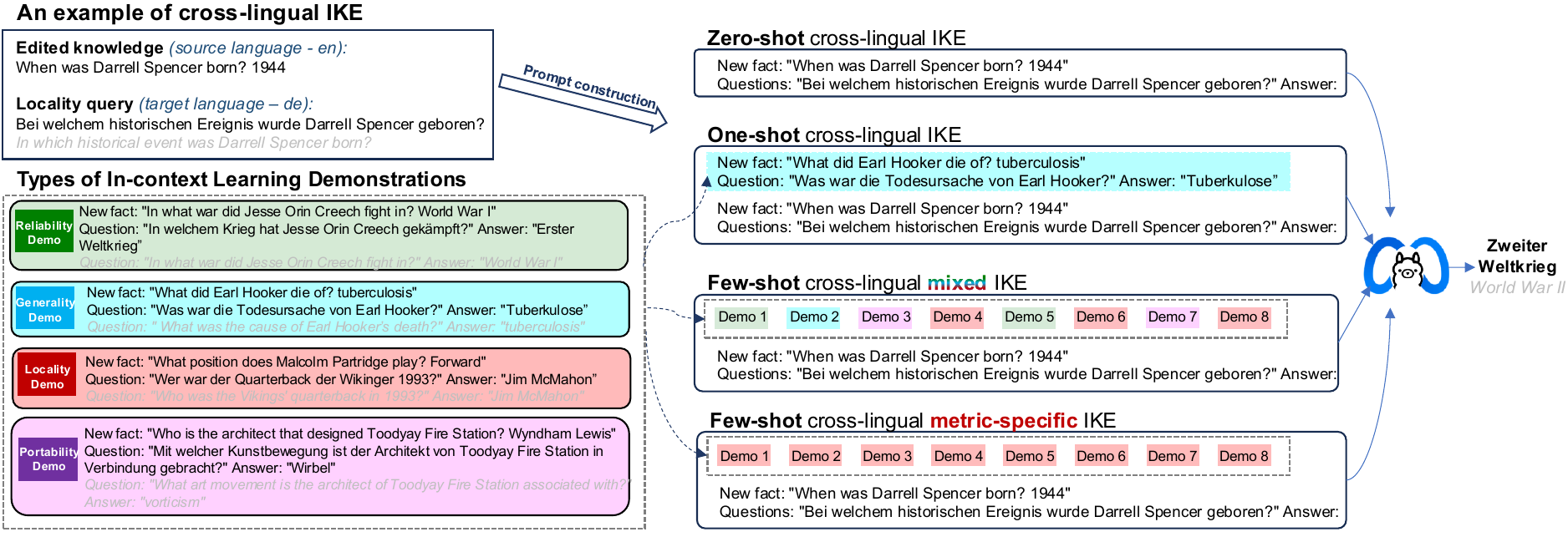}
    \caption{Cross-Lingual IKE setups and demonstration types.}
    \figlabel{ike_pipeline}
\end{figure*}

\section{Related Work}
\seclabel{related_work}
Traditional KE methods are primarily gradient-based. They typically introduce additional trainable parameters, such as MEND~\citep{mitchell2021fast} and SERAC~\citep{mitchell2022memory}---or edit specific
parameters of the original model, as in ROME~\citep{meng2022locating} and MEMIT~\citep{meng2023massediting}. 
However, these methods have high computational demands and are difficult 
to scale. 
Recently, gradient-free methods, such as IKE~\citep{zheng2023can}, MeLLo~\citep{zhong-etal-2023-mquake}, and ICE~\citep{cohen2024evaluating}, have been studied in KE for LLMs. 
Given the multilingual in-context learning capabilities of English-centric LLMs~\citep{lai-etal-2023-chatgpt,nie2024decomposed, zhang-etal-2024-impact}, the potential for cross-lingual KE appears promising.
Recent cross-lingual KE work largely employs gradient-based methods~\citep{xu-etal-2023-language-anisotropic,wang2023cross,beniwal-etal-2024-cross,wei2024mlake}. 
A notable gradient-free work is ReMaKE~\citep{wang2023retrieval}, a cross-lingual retrieval-augmented KE method. However, their method is
specifically applied to a rather special KE scenario---batch edit. In this setting, multiple knowledge pieces, such as the entire knowledge base, are edited simultaneously (\appref{further_discussion}).

\section{BMIKE-53}
BMIKE-53 spans a wide range of knowledge editing perspectives, from artificial to realistic scenarios, and provides a solid foundation for evaluating cross-lingual KE methods. Additionally, with coverage of 53 languages, it stands as the most comprehensive multilingual KE benchmark to date.

\begin{table}[h]
\centering
\scalebox{0.71}{
\begin{tabular}{c|ccc|c}
\toprule
\textbf{Task }       & \textbf{\#Test} & \textbf{Q-Len.} & \textbf{A-Len.} & \textbf{\#Lang.} \\
\midrule
\textbf{zsRE}            & 743    & 9.02         & 2.02           & \multirow{3}{*}{53}       \\
\textbf{CounterFact}    & 1,031  & 5.97          & 1.00           &           \\
\textbf{WFD}          & 784    & 4.71          & 2.55           &     \\
\bottomrule
\end{tabular}}
\caption{Statistics of BMIKE-53. Q/A-Len.: Average Text Length of Query/Answer.}
\tablabel{stat}
\end{table}

\subsection{Datasets}

As shown in \figref{figure1}, BMIKE-53 is constructed from three monolingual KE datasets: \textbf{zsRE}, \textbf{CounterFact}, and \textbf{WikiFactDiff (WFD)}.
Each dataset was selected to represent a distinct perspective of knowledge editing, ensuring the benchmark comprehensively evaluates diverse KE scenarios.

The \textbf{zsRE} dataset, originally introduced by \citet{levy-etal-2017-zero}, was designed for zero-shot relation extraction and later adapted by \citet{de-cao-etal-2021-editing} and \citet{mitchell2021fast} for knowledge editing tasks.
zsRE focuses on regular, well-defined knowledge items, making it an ideal baseline for evaluating the reliability and generality of KE methods. 

The \textbf{CounterFact} dataset, introduced by \citet{meng2022locating}, is designed to evaluate the ability of models to update knowledge with counterfactual (false) facts. Each entry in CounterFact represents a knowledge triple that has been altered to reflect a hypothetical or fabricated scenario.
CounterFact is particularly valuable for assessing the locality of KE methods, as it requires precise updates to counterfactual knowledge without unintended side effects.

The \textbf{WikiFactDiff (WFD)} dataset, introduced by \citet{khodja2024wikifactdiff}, focuses on real-world, temporally recent knowledge updates. Derived from WikiData \citep{vrandevcic2014wikidata}, WFD captures changes to knowledge triples that reflect actual updates in the real world, such as changes in political leadership, scientific discoveries, or other evolving facts.
WikiFactDiff is essential for assessing the real-world applicability of KE methods, as it introduces the challenge of updating models with temporally recent and realistic knowledge changes.

\begin{table*}
\centering
\scalebox{0.7}{
\begin{tabular}{|c|l|cccc|cccc|cccc|} 
\hline
\multirow{2}{*}{\textbf{Model}}
& \multirow{2}{*}{\textbf{Setup}}  & \multicolumn{4}{c|}{\textbf{zsRE}}  & \multicolumn{4}{c|}{\textbf{CounterFact}}           & \multicolumn{4}{c|}{\textbf{WikiFactDiff}}           \\ 
\cline{3-14}
&  & \multicolumn{1}{c}{\textbf{rel}} & \multicolumn{1}{c}{\textbf{gen}} & \multicolumn{1}{c}{\textbf{loc}} & \multicolumn{1}{c|}{\textbf{port}} & \multicolumn{1}{c}{\textbf{rel}} & \multicolumn{1}{c}{\textbf{gen}} & \multicolumn{1}{c}{\textbf{loc}} & \multicolumn{1}{c|}{\textbf{port}} & \multicolumn{1}{c}{\textbf{rel}} & \multicolumn{1}{c}{\textbf{gen}} & \multicolumn{1}{c}{\textbf{loc}} & \multicolumn{1}{c|}{\textbf{port}}  \\ 
\hline
\multirow{4}{*}{\texttt{Llama3.2-3B}} & \textit{zero-shot}       & 50.16    & 49.03    & 5.64     & 5.41      & 43.51    & 40.84    & 7.53     & 5.61      & 58.28    & 57.41    & 6.77     & 3.18       \\ 

             & \textit{one-shot}        & 71.57    & 71.25    & 7.78     & 14.97     & 68.34    & 68.02    & 6.58     & 16.30     & 66.32    & 65.93    & 6.31     & 3.06       \\ 

             & \textit{8-shot (mix)}    & 70.54    & 70.49    & 8.89     & 16.43     & 70.80    & 70.33    & 5.11     & 13.75     & 64.97    & 64.60    & 6.24     & 4.00       \\ 

             & \textit{8-shot (metric)} & 70.94    & 70.91    & 12.23    & 22.97     & 67.57    & 67.14    & 31.61    & 31.20     & 67.77    & 67.48    & 9.14     & 10.72      \\ 
\hline
\multirow{4}{*}{\texttt{Llama3.1-8B}} & \textit{zero-shot}       & 65.53    & 64.09    & 9.76     & 10.05     & 63.01    & 60.59    & 18.68    & 11.16     & 67.84    & 66.40    & 10.04    & 4.15       \\ 

             & \textit{one-shot}        & 75.27    & 74.90    & 13.36    & 20.81     & 71.92    & 71.29    & 12.66    & 21.93     & 70.53    & 69.84    & 7.80     & 4.15       \\ 

             & \textit{8-shot (mix)}    & 74.29    & 74.00    & 15.46    & 25.18     & 75.15    & 74.42    & 11.40    & 23.74     & 68.57    & 67.86    & 8.27     & 8.87       \\ 

             & \textit{8-shot (metric)} & 74.86    & 74.79    & 16.15    & 32.86     & 73.88    & 73.19    & 47.55    & 41.17     & 71.98    & 71.34    & 13.84    & 14.58      \\
\hline
\end{tabular}}
\caption{Main Results. Average cross-lingual IKE performance across 52 languages (F1-score).}
\tablabel{main_result}
\end{table*}

\subsection{Benchmark Construction}

The construction of the BMIKE-53 benchmark involves three key steps: unifying data formats, multilingual expansion, and quality control. These steps ensure that the benchmark is consistent, multilingual, and of high quality, enabling robust evaluation of cross-lingual knowledge editing (KE) methods.

\paragraph{Unifying Data Formats}
To ensure consistency across the three datasets, we standardized their formats into a unified structure to create a cohesive English base dataset.
Each data item in the unified format includes an edited knowledge item and four types of test queries: reliability, generality, locality, and portability.
The portability queries for zsRE and CounterFact were adopted from the work of \citet{yao2023editing}, while for WFD, we extracted a knowledge graph from the original WFD dataset and performed one-hop knowledge reasoning within the graph to generate portability queries. 
All data items are stored in a JSON format, with a consistent data structure across the three datasets. This unified format facilitates seamless integration into the benchmark and enables efficient LLM-assisted translation in the multilingual expansion process.

\paragraph{Structured Multilingual Expansion with LLMs}
To create a multilingual benchmark for cross-lingual KE tasks, we expanded the English base data into 52 target languages using LLM-assisted structural translation. The language coverage is adopted from similar multilingual datasets like MLAMA~\citep{kassner2021multilingual} and BMLAMA~\citep{qi2023cross}. 
App. \ref{lang_cov} provides details regarding the language coverage.

We employed the GPT-4o model\footnote{\texttt{gpt-4o-2024-08-06}} via the OpenAI API for the multilingual expansion. The translation process was guided by a structured prompt template (see Appendix A.3), which ensured that the JSON format and structure of the data items were preserved during translation. Specifically:
Only the text values within the JSON structure were translated, while the key names remained unchanged.
The prompt explicitly instructed the model to output the translated data in plain text, adhering to the original JSON format. We compared several machine translation tools, including NLLB-200 and Google Translate API, but found that LLM-assisted translation was more appropriate for our use case due to its accuracy and flexibility in handling structured data formats (\appref{translation_prompt}).

\paragraph{Quality Control}
To ensure the quality of the multilingual expansion, we implemented a rigorous quality control process involving qualitative evaluation and quantitative analysis.
We conducted a manual review of sampled sentences by native speakers of selected languages, then we used back-translation techniques to provide an overall assessment of translation quality. Specifically, each translated sentence was back-translated into English, and the BLEU score and semantic similarity were calculated. \tabref{translation_quality} shows the results of translation quality control via the back-translation.

\section{Experiments}

The primary goal of this work is to extensively investigate the performance of cross-lingual in-context knowledge editing (IKE). Using the proposed benchmark BMIKE-53, we aim to explore the factors influencing cross-lingual IKE performance, identify performance tendencies, and analyze variations in cross-lingual behavior across different languages and query types. To achieve this, we first formally define the cross-lingual IKE task and its evaluation framework. We then introduce the different IKE setups and strategies explored in our experiments

\begin{figure*}[t]
    \centering
    \includegraphics[width=\linewidth]{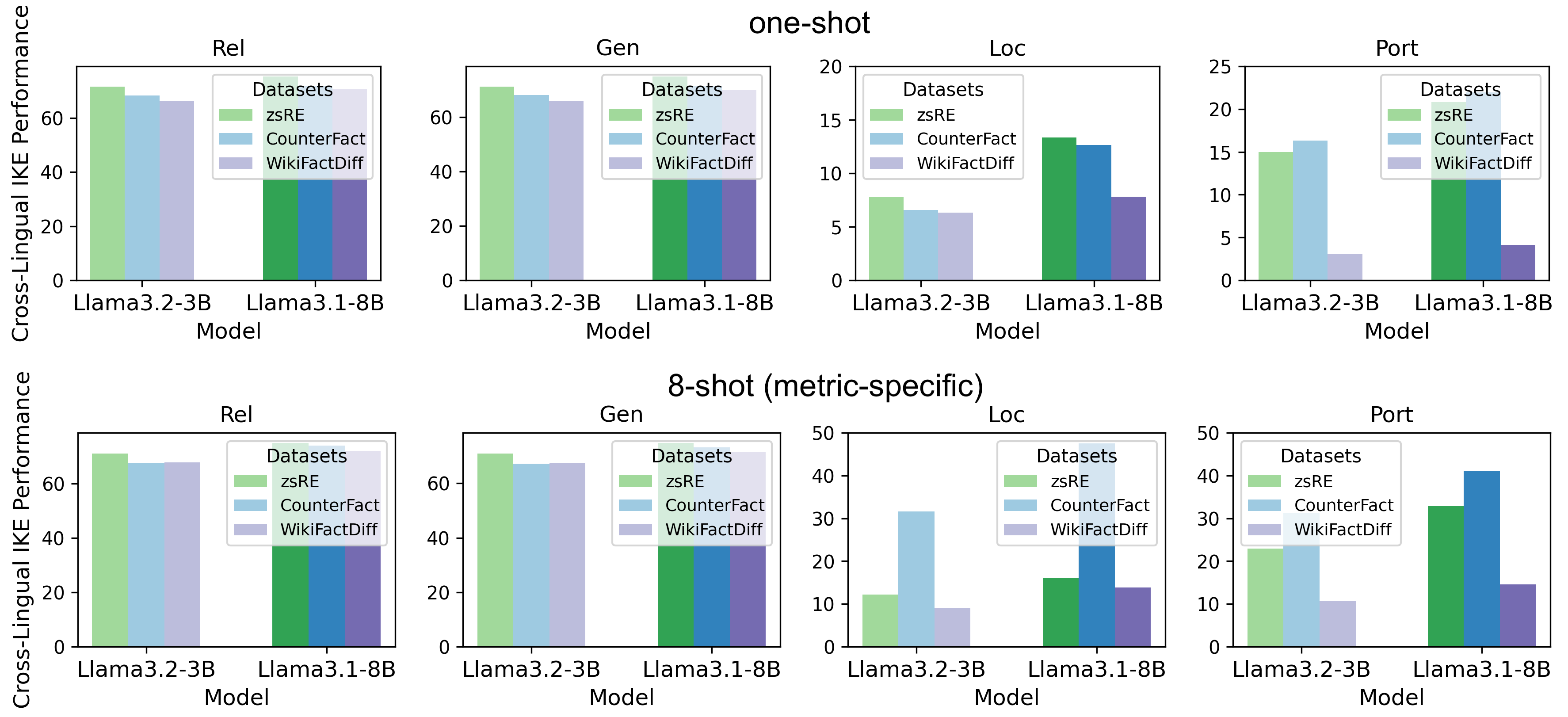}
    \caption{Average cross-lingual IKE performance across languages (F1-score).}
    \figlabel{bar_plots}
\end{figure*}

\subsection{Task: Cross-Lingual In-Context Knowledge Editing}
The Cross-Lingual In-Context Knowledge Editing (IKE) task evaluates a language model's ability to incorporate new knowledge (a fact) in one language and apply it across multiple languages while preserving unrelated knowledge. This task leverages in-context learning (ICL) to guide the model in editing and applying knowledge through demonstrations. Below, we formally define the task and the four types of cross-lingual queries used to evaluate the model's performance.

\paragraph{Task Definition} 
Given a language model $\mathcal{M}$; a new fact represented as a query-answer pair $f=(x_s^*, y_s^*)$ in the source language $s$, where $x_s^*$ is the query and $y_s^*$ is the corresponding answer; a set $\mathcal{X}_s^*$, which contains $x_s^*$ and other semantically equivalent queries in the source language; the translations of $\mathcal{X}_s^*$ into a target language $t$, denoted as $\mathcal{X}_t^* = \{I_t(x_s): x_s\in \mathcal{X}_s^*\}$ where $I^t(\cdot)$ is a translator mapping source language queries to their target language counterparts.
The task involves evaluating the model's response to a target language query $x_t$ after incorporating the new fact $f$. Specifically, the model assigns a probability ${P}_{\mathcal{M}}(y|x_t, f)$ to an answer $y$ given the query $x_t$ and the fact $f$. The predicted answer is defined as $Pred(x_t, f)=\argmax_{y}\mathcal{P}_{\mathcal{M}}(y|x_t, f)$.
The goal is for the model to predict the correct translation of the fact answer, $I^t(y_s^*)$, when the query $x_t$ is semantically equivalent to the fact query $x_s^*$, and to preserve the original knowledge and predict the correct answer $y_t$ for unrelated queries, ensuring no unintended inference from the knowledge editing process.

\paragraph{Cross-Lingual Query Types}
To comprehensively evaluate the model's cross-lingual knowledge editing capabilities, we define four types of target language queries, each testing a specific aspect of the task. The model should reliably apply the new fact to the exact translation of the original query. A \textbf{\textit{reliability}} query $x_t$ is defined as $x_t=I_t(x_s^*)$. 
\textbf{\textit{Generality}} queries test whether the model can generalize the new fact to other semantically equivalent queries in the target language that differ in phrasing or structure. A generality query $x_t$ is defined as $x_t\in\mathcal{X}_t^*\backslash\{I_t(x_s^*)\}$.
The expected answer for the reliability and generality query is $Pred(x_t, f)=I_t(y_s^*)$.
\textbf{\textit{Portability}} queries evaluate whether the model can apply the new fact to related but contextually different queries in the target language. These queries are derived through one-hop knowledge reasoning from the original fact. Let $\mathcal{X}^*_{s\mbox{,1-hop}}$ denote the one-hop query set in the source language, which includes $x_s^*$ and queries influenced by $x_s^*$ through one-hop reasoning in a knowledge graph. The corresponding target language set is $\mathcal{X}^*_{t\mbox{,1-hop}}=\{I_t(x_s^*):x_s\in\mathcal{X}^*_{s\mbox{,1-hop}}\}$. A portability query $x_t$ is defined as $x_t\in \mathcal{X}^*_{t\mbox{,1-hop}}$.
\textbf{\textit{Locality}} queries test whether the model can preserve unrelated knowledge while incorporating the new fact. A locality query $x_t$ is defined as $x_t\notin \mathcal{X}^*_{t\mbox{,1-hop}}$. The expected answer is $Pred(x_t, f)=y_t$.

\begin{figure*}[t]
    \centering
    \includegraphics[width=\linewidth]{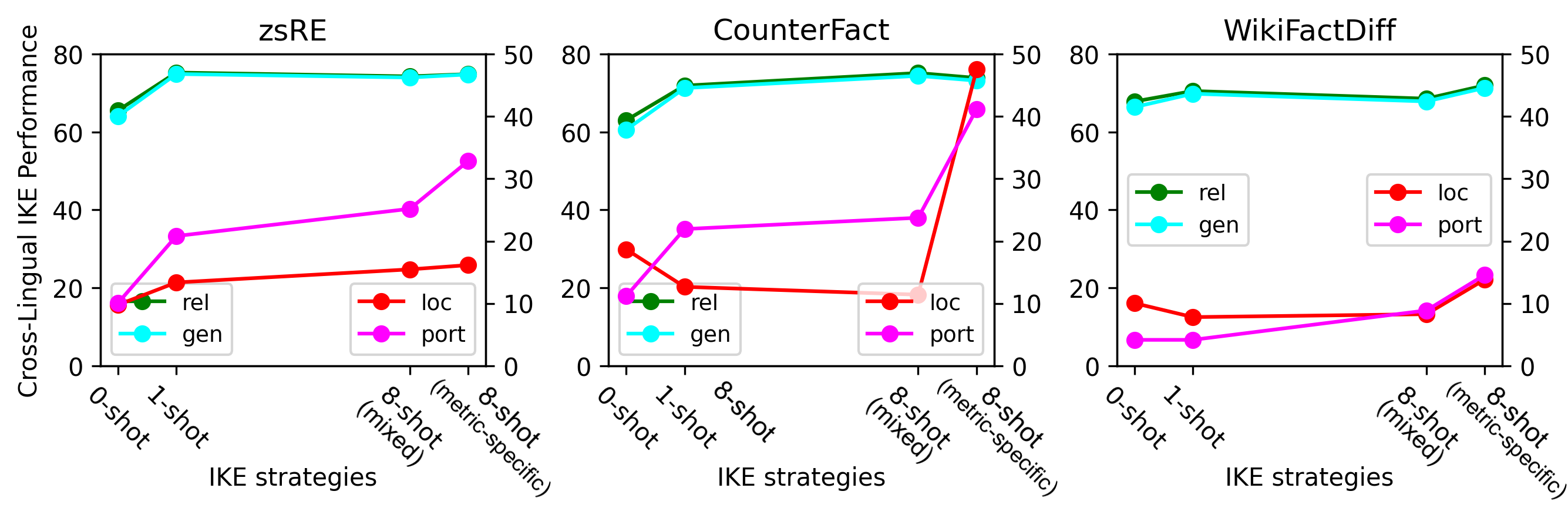}
    \caption{Average cross-lingual IKE performance of Llama3.1-8B across 52 languages (F1-score). }
    \figlabel{line_plot}
\end{figure*}

\subsection{Cross-Lingual IKE Setup}

\paragraph{IKE Demonstrations}
In the context of in-context learning (ICL), demonstrations are examples provided in the input prompt to guide the model's behavior. For the IKE task, the demonstrations are designed to teach the model how to perform cross-lingual knowledge editing. Formally, the set of demonstrations is defined as $C=\{c_1, \cdots, c_k\}$, where each demonstration $c_i$ consists of a new fact $f'=(x_s', y_s')$ in the source language, a query $x_t'$ in the target language, and the correct answer $y_t'$ in the target language.
As an ICL-based KE method, IKE uses demonstrations to guide the model in learning the relationships between the source language fact and the target language queries. As illustrated in \figref{ike_pipeline}, the demonstrations are designed to reflect the four query types, enabling the model to learn the appropriate behavior for each type. By including examples of cross-lingual queries and their correct answers, the demonstrations help the model generalize the knowledge editing process across languages.

\paragraph{IKE Setup}
As illustrated in \figref{ike_pipeline}, we evaluate cross-lingual IKE under four distinct setups: zero-shot, one-shot, few-shot mixed, and few-shot metric-specific. 
In \textbf{\textit{zero-shot}} cross-lingual IKE, the model performs cross-lingual knowledge editing without any demonstrations, relying solely on its pre-trained capabilities. In a \textbf{\textit{one-shot}} setup, a single randomly selected demonstration is provided to familiarize the model with the task format. This setup is designed to give the model an overview of the task without significantly aiding its ability to complete the task. \textbf{\textit{Few-shot mixed}} cross-lingual IKE includes eight demonstrations of mixed types. The goal is to teach the model cross-lingual knowledge editing by exposing it to diverse query types. To enhance performance on specific query types, the \textbf{\textit{few-shot metric-specific}} variation provides eight demonstrations of the same query type as the test target.

\subsection{Experimental Setting}
\seclabel{setting}
We conduct experiments using two multilingual LLMs: Llama3.2-3B and Llama3.1-8B.  
These models were selected for their multilingual capabilities and represent different model sizes, allowing us to analyze the impact of model scale on cross-lingual IKE performance.
To measure the cross-lingual IKE performance, we
compare the predicted answers with the ground-truth answers using the \textbf{F1} score and Exact Match (\textbf{EM}) metrics, consistent with prior work~\citep{wang2023cross,wang2023retrieval}.
The implementation details are provided in \appref{setting}.

\begin{figure}[h]
    \centering
    \includegraphics[width=\linewidth]{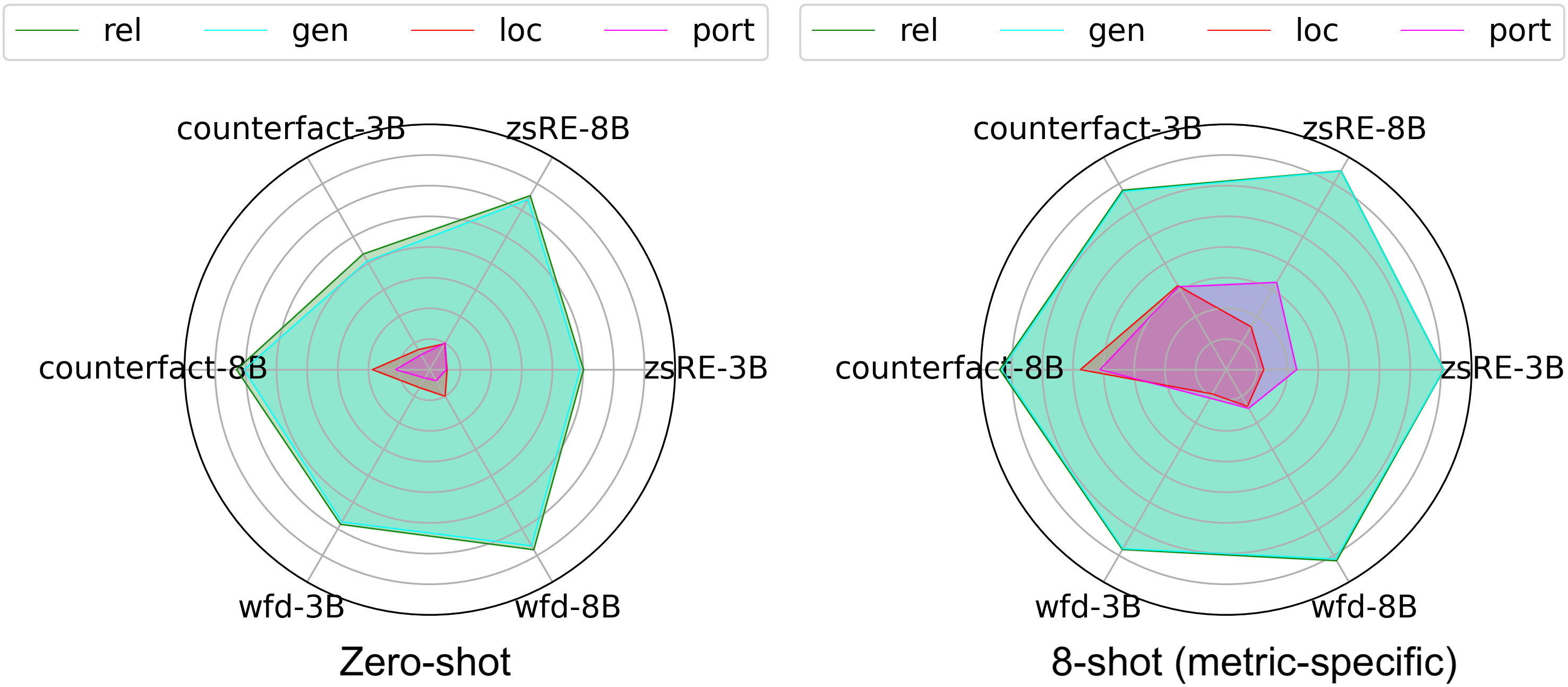}
    \caption{Average cross-lingual IKE performance across 52 languages (F1-score).}
    \figlabel{radar}
\end{figure}

\section{Multidimensional Analysis of Cross-Lingual IKE}
This section provides a multidimensional analysis of cross-lingual IKE performance, focusing on the effects of model size, dataset-specific performance, query type variations, and IKE setup strategies. Using the BMIKE-53 benchmark, we aim to uncover key insights into how these factors influence IKE performance and cross-lingual variations.

\begin{table*}[t]
    \centering
    \includegraphics[width=\linewidth]{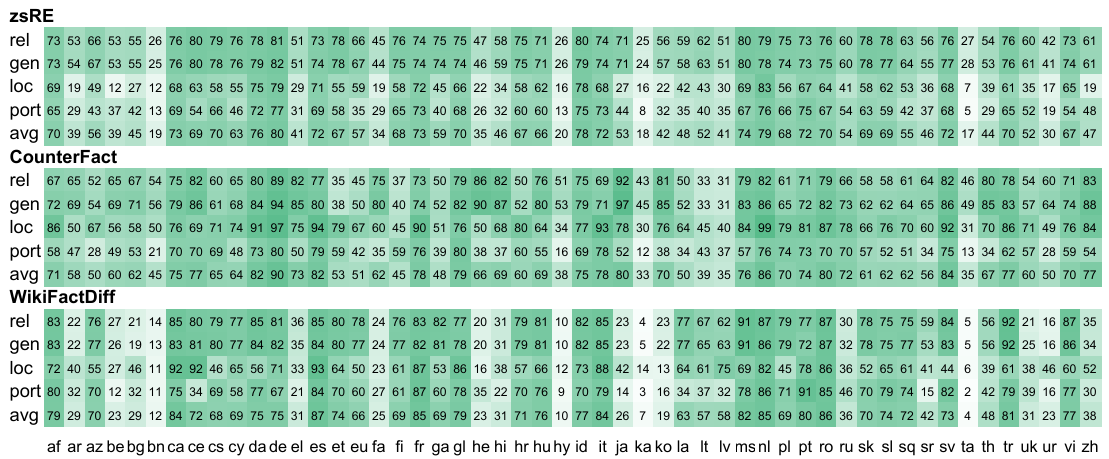}
    \caption{Cross-lingual IKE performance of Llama3.1-8B in 52 languages under the 8-shot metric-specific setup.}
    \tablabel{multi_results}
\end{table*}

\begin{figure*}
    \centering
    \includegraphics[width=\linewidth]{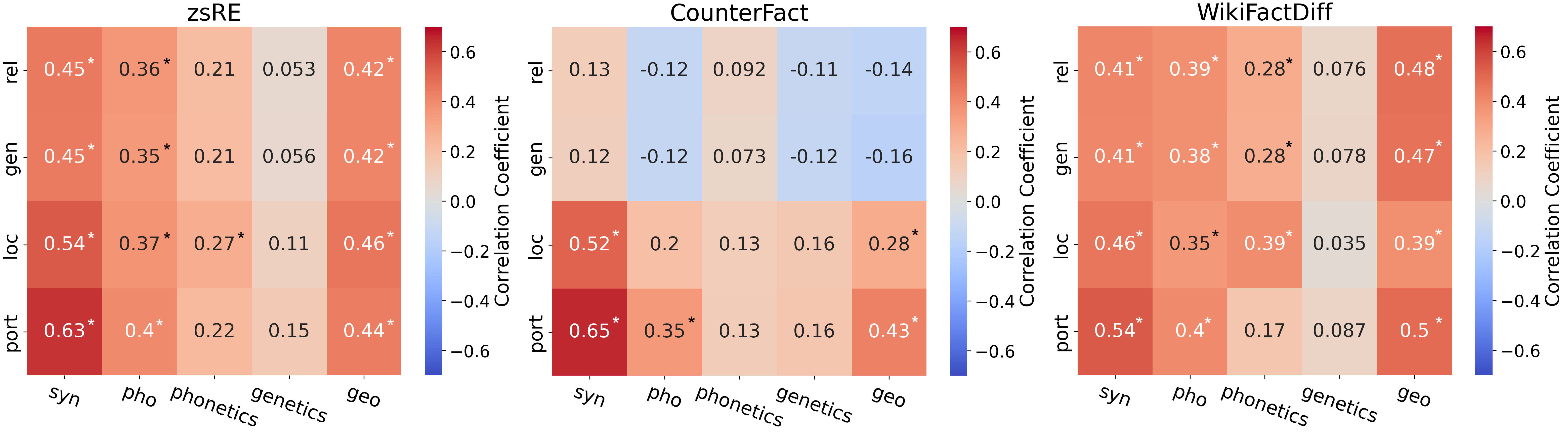}
    \caption{Correlation between linguistic properties and IKE performance of 52 languages. Experimental results of Llama3.1-8B under the 8-shot metric-specific setup. Significant correlations are marked with $*$ ($p<0.05$).}
    \figlabel{heatmap}
\end{figure*}

\subsection{Effects of Model Scale}
As shown in \figref{bar_plots}, larger models demonstrate superior cross-lingual IKE capabilities across all experimental configurations. 
Notably, larger models show particular advantages in handling queries requiring complex multilingual reasoning and knowledge preservation, evidenced by the performance gap of locality (loc) and portability (port) queries.

\subsection{Dataset-Specific Performance Patterns}
We observe substantial cross-dataset variance in editing efficacy, especially with loc and port queries, indicative of inherent dataset complexity differences. While WikiFactDiff (WFD) achieves comparable reliability (rel) and generality (gen) scores to zsRE and CounterFact, it shows the lowest port performance. 
This discrepancy could be attributed to the real-world nature of WFD,  where all knowledge--both the original and updated facts--is temporally recent.
Portability queries in WFD require the model to reason over a second-order knowledge chain, where the correct answer depends on understanding the relationship between the updated fact and its broader context. 
CounterFact, on the other hand, benefits the most from metric-specific demonstrations, particularly for locality and portability. The compositional nature of CounterFact's fabricated facts allows the model to leverage targeted demonstrations effectively.

\subsection{Query-Type Sensitivity}
The four query types exhibit divergent response patterns to IKE strategies. \figref{radar} shows that rel and gen achieve near-parity across setups, especially in 8-shot metric-specific IKE, suggesting models effectively align cross-lingual surface forms.
In contrast, loc and port perform substantially worse across datasets and setups. However, port demonstrates greater sensitivity to metric-specific demonstrations than loc. Port benefits from metric-specific demonstrations with more pronounced improvements, especially in datasets like zsRE and WFD.

\subsection{Impact of Demonstration Strategies}
The comparison of zero-shot, one-shot, few-shot mixed, and few-shot metric-specific setups reveals the importance of demonstration quality and quantity in cross-lingual IKE. \figref{line_plot} illustrates how the performance of \texttt{Llama3.1-8B} evolves across setups for each query type and dataset.
In the zero-shot setup, models rely solely on their pre-trained capabilities, resulting in limited performance. 
While one-shot setups provide modest improvements by familiarizing the model with the task format, they fail to address the challenges of loc.
A single, randomly selected demonstration often confuses the model, particularly when the demonstration type does not align with the target query type. 
For loc queries, when encountering rel or gen demonstration types, this misalignment can lead the model to incorrectly repeat the edited knowledge in the target language, further degrading performance.
Adding more demonstrations in the few-shot mixed setup yields limited performance gains, particularly for loc queries. However, when demonstrations are tailored to the specific query type being tested, as in the 8-shot metric-specific setup, notable gains are observed. This setup achieves the highest scores for loc and portability, as shown in \figref{line_plot}. These findings underscore the importance of demonstration quality and specificity in maximizing the benefits of in-context learning for cross-lingual knowledge editing.

\begin{figure}[h]
    \centering
    \includegraphics[width=.97\linewidth]{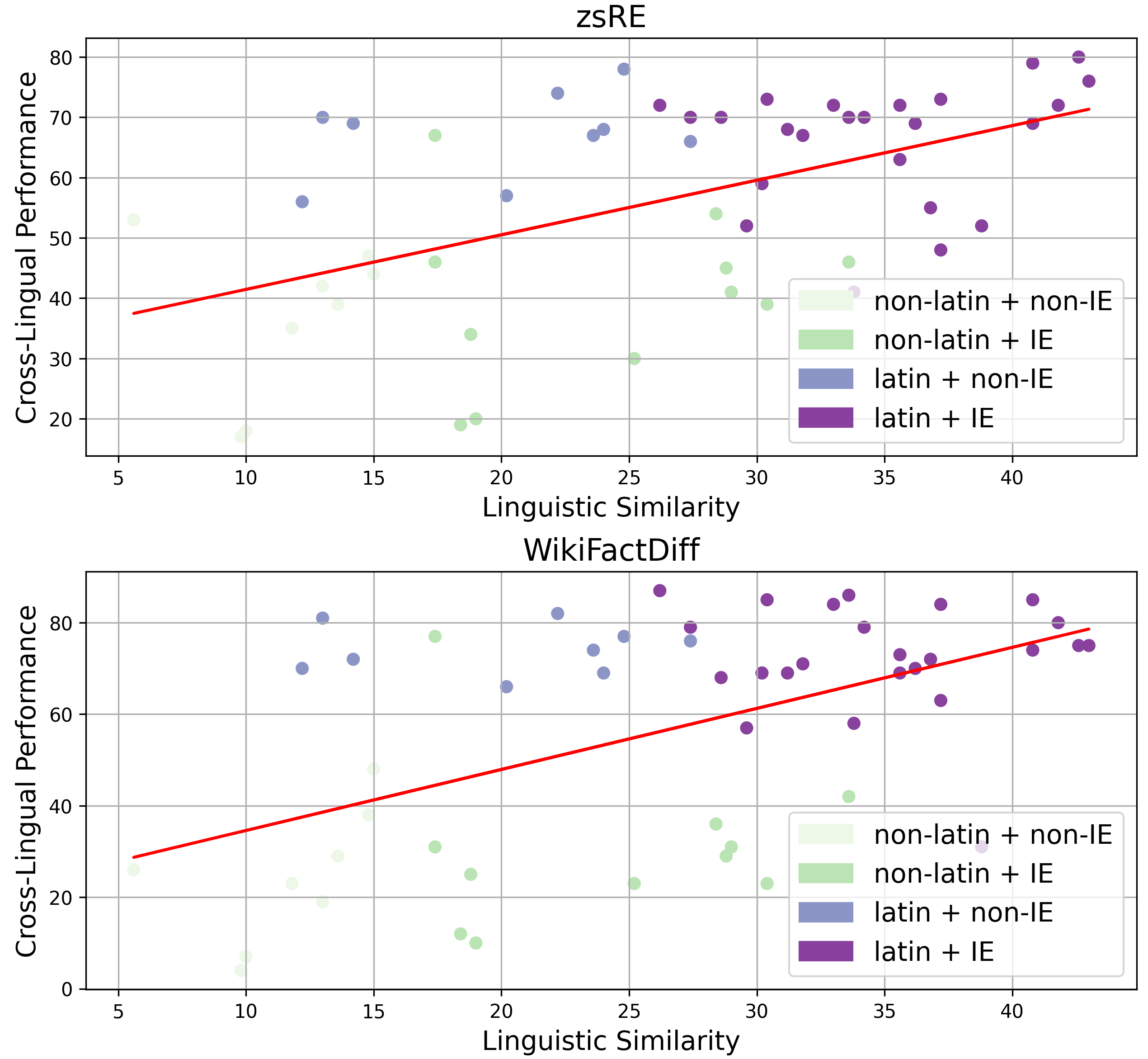}
    \caption{Panorama of per-language performance.}
    \figlabel{scatter}
\end{figure}

\section{Language Performance Variance in Cross-Lingual IKE}

To analyze cross-lingual performance variation, we use a normalized metric that uses the exact match (EM) of English as a reference to highlight cross-lingual transfer differences. Specifically, we calculate the ratio of each target language's EM performance to English. As is visually evident in \tabref{multi_results}, some languages perform particularly poorly, while others achieve results closer to English, motivating further investigation into the factors influencing these differences.

\subsection{Correlation Between Language Properties and Performance}
We conducted a correlation analysis between language properties (derived using Lang2Vec~\citep{littell-etal-2017-uriel}, details in Appendix) and query-type performance across datasets.
As revealed in \figref{heatmap}, syntactic, phonological, and geographic similarities with English positively correlate with performance, particularly for loc and port queries However, two notable exceptions emerge: rel and gen queries in CounterFact show no significant correlation with language properties, likely because these query types achieve uniformly high performance across all languages. Additionally, genetic similarity (language family) shows no meaningful correlation across datasets.

\subsection{Impact of Script and Language Family}
To further explore the role of script and language family, we grouped languages into four clusters based on script type (Latin vs. non-Latin) and language family (Indo-European vs. non-Indo-European). \figref{scatter} reveals that script type plays a more critical role than language family. Non-Latin languages, regardless of their family, perform worse than Latin-script languages. This trend is consistent across datasets and query types, as further supported by \figref{box}. For example, the non-Latin + non-IE and non-Latin + IE groups exhibit similar performance, both significantly lower than the Latin + IE and Latin + non-IE groups. 

\begin{figure}[h]
    \centering
    \includegraphics[width=\linewidth]{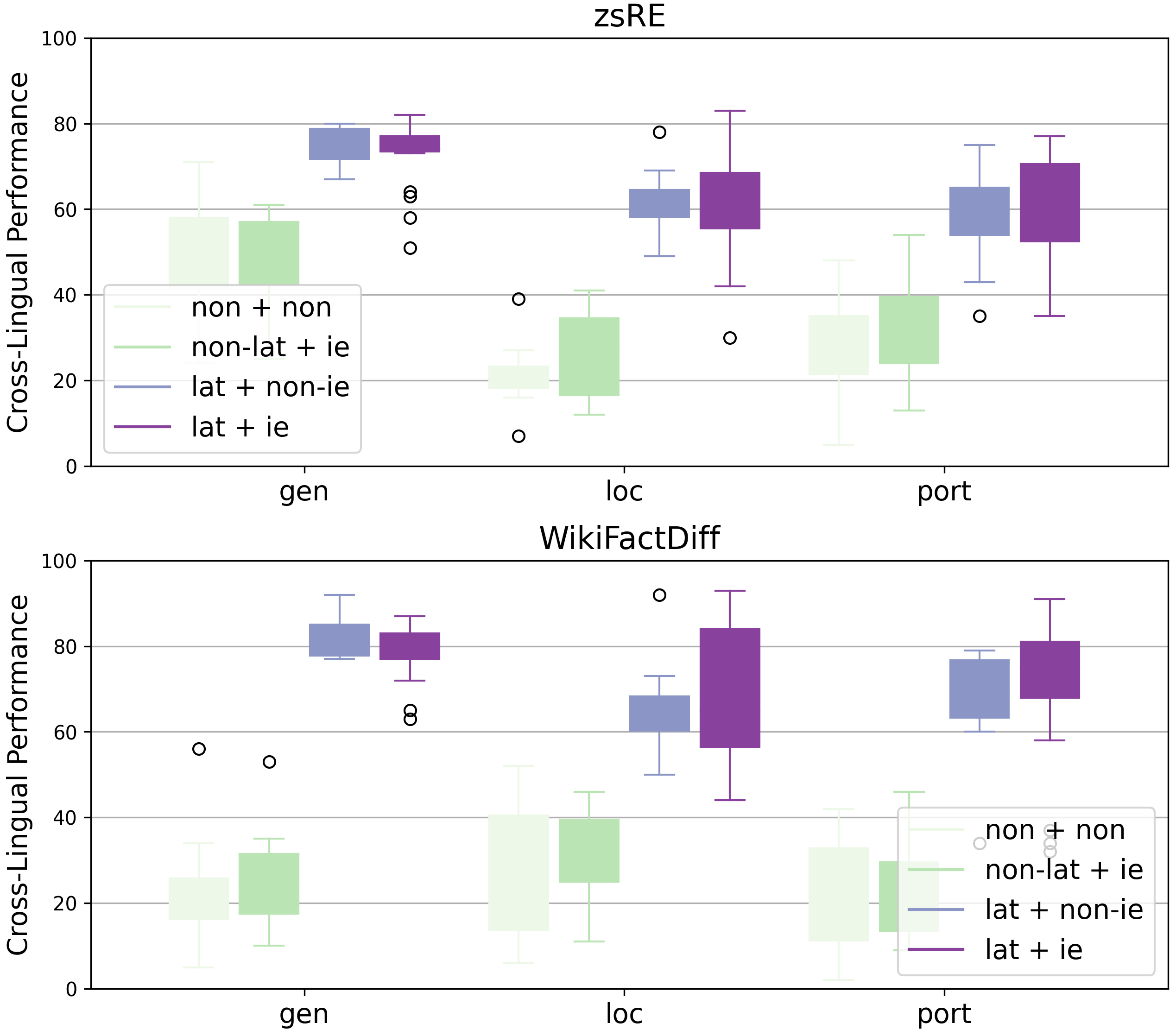}
    \caption{IE: Indo-European language family.}
    \figlabel{box}
\end{figure}

\subsection{Language Confusion and Script Effects}
A qualitative error analysis highlights language confusion as a key factor explaining why script type matters. Language confusion occurs when the model generates answers in English instead of the target language \citep{marchisio-etal-2024-understanding}, even when explicitly instructed to use the target language. This issue is particularly prevalent in code-switched prompts and disproportionately affects non-Latin languages. For example, while the model may produce the correct answer in English, it is considered a failure if the output is not in the target language. 
\tabref{case} highlights instances of language confusion that occur when queries are presented in Chinese.
This problem is exacerbated for non-Latin languages, as their distinct scripts reduce the likelihood of overlapping writing forms between English and the target language, further degrading performance.

\begin{table}[t]
    \centering
    \includegraphics[width=\linewidth]{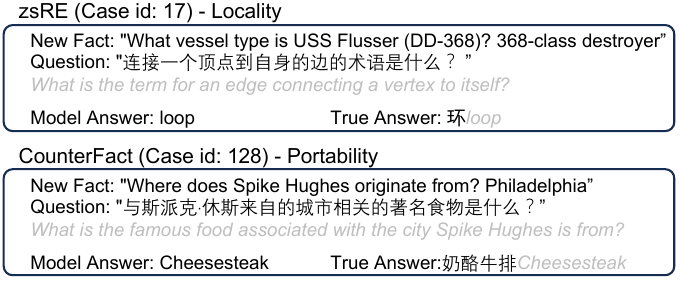}
    \caption{Examples of language confusion from the 8-shot metric-specific setup with Llama3.1-8B.}
    \tablabel{case}
\end{table}

\section{Conclusion}

We present BMIKE-53, a multilingual KE benchmark, and leverage it to investigate the potential of gradient-free in-context learning methods for cross-lingual knowledge editing. Our experiments demonstrate that tailored demonstration strategies significantly enhance KE performance, with metric-specific demonstrations improving locality and portability. Additionally, linguistic properties, particularly script type, strongly influence cross-lingual knowledge transfer. We hope that BMIKE-53 will inspire further research in multilingual KE, advancing the understanding and capabilities of LLMs across diverse linguistic contexts.

\section*{Limitation}

This research focuses on the application of gradient-free in-context learning (ICL) methods for cross-lingual knowledge editing (KE) using BMIKE-53, the most comprehensive multilingual KE benchmark to date. While our study highlights the effectiveness of ICL-based methods and tailored demonstration strategies, certain areas remain for further exploration. For instance, gradient-based KE methods, which are computationally intensive, were explored very deeply in this investigation as our primary focus was on the efficiency and practicality of gradient-free approaches.
Additionally, while BMIKE-53 provides valuable insights into the challenges of linguistic diversity, future work could aim to further optimize performance for non-Latin languages and address language confusion issues. These extensions could deepen our understanding of cross-lingual KE and enhance the applicability of BMIKE-53 as a resource for advancing multilingual LLM research.

\section*{Ethic Statement}
This research was conducted in accordance with the ACM Code of Ethics. 
The datasets used in this paper are publicly available. 
We 
do not intend to share any Personally Identifiable Data with this paper.
Our project may raise awareness of these low-resource languages in the computational linguistics community.
Artificial intelligence (AI) tools were utilized exclusively for language refinement during the preparation of this manuscript. 
The use of AI tools is disclosed here in accordance with current ethical guidelines and publishing policies.

\section*{Acknowledgments}
We want to thank the anonymous reviewers for their valuable feedback and constructive suggestions, which have helped to improve the quality of this paper.
This work was supported by the German Research Foundation (DFG) under grant SCHU 2246/14-1, Munich Center for Machine Learning (MCML), and China Scholarship Council (CSC).

\bibliography{custom}

\appendix

\section{BMIKE-53 Details}
\applabel{sec:appendix}

\subsection{Language Coverage}
\label{lang_cov}

BMIKE-53 encompasses a total of 53 languages. A comprehensive list of these languages can be found in~\tabref{lang_info}. Additionally, the table outlines the linguistic feature similarities between each target language and English.

\subsection{Data Entry Example}
\applabel{data_example}
\figref{example} shows the data item examples of BMIKE-53 for all three datasets.

\subsection{LLM-Assisted Translation}
\applabel{translation_prompt}
\figref{prompt_template} shows the prompt template used for our LLM-assisted translation.
\paragraph{Why LLM-Assisted Translation?}
We compared several machine translation tools, including NLLB-200 and Google Translate API, but found that LLM-assisted translation offered superior performance in terms of:
\begin{itemize}
    \item Accuracy: LLMs demonstrated better handling of complex linguistic structures and domain-specific terminology.
    \item Flexibility: LLMs provided greater adaptability for processing structured data formats like JSON.
    \item Consistency: The structured translation process ensured that the multilingual data remained aligned with the original English data.
\end{itemize}

By leveraging LLM-assisted translation, we ensured that the multilingual benchmark maintained high linguistic quality and structural integrity.

\begin{figure}
    \centering

    \begin{tcolorbox}

\textbf{system:}

You are an intelligent multilingual translation assistant that can structurally translate English text in a fixed data format into 52 different languages.\\

\textbf{user:}\\
Translate the following JSON data item from English to \texttt{\{target language\}}. Keep the original JSON format and structure, translating only the text in the values while keeping the key names unchanged. Output only in plain text without additional formatting text. Use double quotes for key and value names and add the escape character for quote marks in the text of value:
\texttt{\{json\_data\_item\}}

\end{tcolorbox}
    
    \caption{Prompt template for \texttt{GPT-4o} as translation assistant.}
    \figlabel{prompt_template}
\end{figure}

\subsection{Translation Quality Control}
We used back-translation techniques to assess the quality of the translations. Specifically, each translated sentence was back-translated into English. We calculated the BLEU score and semantic similarity between the original and the back-translated English text.
BLEU Score measures the formal similarity between the original and back-translated sentences.
Semantic Similarity evaluates the semantic alignment between the original and back-translated sentences using cosine similarity in a sentence embedding space.
The results of the back-translation evaluation are shown in \tabref{translation_quality}.
\begin{table}[h]
\centering
\scalebox{0.8}{
\begin{tabular}{c|cc|cc|cc} 
\toprule
& \multicolumn{2}{c|}{\textbf{zsRE}} & \multicolumn{2}{c|}{\textbf{CounterFact}} & \multicolumn{2}{c}{\textbf{WFD}}\\
\hline
 \textbf{Lang.} &\textbf{BLEU}          & \textbf{Sim.}     &\textbf{BLEU}          & \textbf{Sim.}&\textbf{BLEU}          & \textbf{Sim.}      \\ 
\hline     
      
\textbf{es}          & 0.81           & 0.94 & 0.82           & 0.90   & 0.81           & 0.90 \\ 
{\textbf{vi}}           & 0.82           & 0.93     & 0.77     & 0.91 & 0.78  & 0.90\\ 
\textbf{ru}           & 0.78           & 0.91  & 0.71   & 0.87 & 0.72  & 0.87       \\ 
\textbf{zh}           & 0.78           & 0.89  & 0.76  & 0.85   & 0.76   & 0.85     \\ 
\textbf{de}           & 0.84 & 0.93    & 0.82      & 0.92  & 0.82           & 0.92  \\ 

\bottomrule
\end{tabular}}
    \caption{Results of Translation Quality Control via Back-Translation.}
    \tablabel{translation_quality}
\end{table}

\section{BMIKE-53 with Other Baseline Methods}
\applabel{other_baselines}
The primary purpose of this paper is to provide a basic gradient-free method for cross-lingual KE. However, it is helpful for a better understanding of our gradient-free method to compare it with other existing baseline methods, including gradient-based methods. Therefore, we also experimented on BMIKE-53 with commonly used baseline methods, including fine-tuning (FT), LoRA~\citep{hulora}, ROME~\citep{meng2022locating}, and KN~\citep{dai-etal-2022-knowledge} in selected languages. We use the EasyEdit framework~\citep{wang-etal-2024-easyedit} to conduct the baseline experiments. \tabref{baseline_results} shows the baseline method results.

\begin{table}[!t]
    \centering
    \scalebox{.8}{
    \begin{tabular}{c|c}
    \toprule
       \textbf{Parameter}  & \textbf{Value} \\
       \bottomrule
       \multirow{2}{*}{Model}  & \multirow{2}{*}{\parbox{5cm}{\texttt{meta-llama/Llama-3.1-8B}, \newline \texttt{meta-llama/Llama-3.2-3B}}} \\
       & \\
       \hline
       Max. length & 4096 \\
       \hline
       Num. of demonstration & 8 \\
       \hline
       \multirow{2}{*}{Type of demonstration} & \multirow{2}{*}{\parbox{5cm}{Reliability, Generality, \newline Locality, Portability}} \\
       & \\
       \hline
       Proportion of demo. & 1:3:2:2 \\
       \hline
       GPU Type & NVIDIA A100-SXM4-80GB \\
       \hline
       Number of GPU & 4 \\
       \hline
       Running hours & 72 \\
       \bottomrule
    \end{tabular}}
    \caption{Experimental Implementation Details.}
    \tablabel{setting}
\end{table}

\begin{table*}[t]
    \centering
    \scalebox{.85}{
    \begin{tabular}{c|c|cccccccccc}
    \hline
          \textbf{Query} &\textbf{Baseline} & \textbf{ar}    & \textbf{de}    & \textbf{fa}    & \textbf{fr}  & \textbf{he}    & \textbf{hu}    & \textbf{ja}    & \textbf{ru}  & \textbf{tr}    & \textbf{zh}\\
    \hline
\multirow{4}{*}{\textbf{Reliability}}  & \textbf{FT}       & 24.9                    & 22.4                    & 20.5                    & 19.6                    & 26.0                    & 35.0                    & 34.4                    & 17.0                    & 24.1                    & 23.4                     \\ 

                      & \textbf{LoRA}     & 47.1                    & 46.0                    & 47.0                    & 44.8                    & 43.9                    & 44.1                    & 53.1                    & 35.4                    & 41.3                    & 47.9                     \\ 

                      & \textbf{ROME}     & 34.9                    & 36.3                    & 35.8                    & 33.7                    & 25.8                    & 32.2                    & 42.6                    & 24.8                    & 30.7                    & 38.0                     \\ 

                      & \textbf{KN}       & 20.1                    & 3.7                     & 24.0                    & 3.7                     & 15.8                    & 5.1                     & 28.0                    & 8.5                     & 5.7                     & 23.2                     \\ 
\hline
\multirow{4}{*}{\textbf{Generality}}  & \textbf{FT}       & 25.6                    & 23.6                    & 21.7                    & 20.8                    & 25.9                    & 34.5                    & 34.0                    & 17.1                    & 25.2                    & 24.5                     \\ 

                      & \textbf{LoRA}     & 44.6                    & 45.4                    & 46.7                    & 44.4                    & 42.7                    & 42.8                    & 52.7                    & 34.8                    & 41.0                    & 47.0                     \\ 

                      & \textbf{ROME}     & 33.7                    & 35.1                    & 36.6                    & 32.7                    & 25.3                    & 30.4                    & 43.0                    & 25.4                    & 31.3                    & 38.8                     \\ 

                      & \textbf{KN}       & 20.7                    & 3.5                     & 24.8                    & 3.7                     & 16.1                    & 5.5                     & 28.2                    & 8.4                     & 5.8                     & 24.1                     \\ 
\hline
\multirow{4}{*}{\textbf{Locality}}  & \textbf{FT }      & 37.5                    & 18.7                    & 31.6                    & 15.1                    & 41.0                    & 35.6                    & 39.9                    & 32.0                    & 25.7                    & 39.8                     \\ 

                      & \textbf{LoRA}     & 38.1                    & 26.3                    & 37.4                    & 26.1                    & 41.0                    & 35.1                    & 42.3                    & 32.4                    & 29.6                    & 45.5                     \\ 

                      & \textbf{ROME}     & 78.9                    & 75.8                    & 79.8                    & 78.3                    & 83.9                    & 87.3                    & 80.8                    & 77.7                    & 83.2                    & 84.6                     \\ 
                      & \textbf{KN}       & 51.3                    & 60.2                    & 54.6                    & 57.5                    & 62.8                    & 64.7                    & 56.8                    & 54.7                    & 60.3                    & 61.1                     \\ 
\hline
\multirow{4}{*}{\textbf{Portability}} & \textbf{FT}       & 32.7                    & 16.8                    & 28.9                    & 14.1                    & 34.7                    & 18.5                    & 35.1                    & 19.6                    & 19.0                    & 27.6                     \\ 

                      & \textbf{LoRA}     & 42.9                    & 27.2                    & 42.8                    & 24.0                    & 40.4                    & 23.7                    & 44.4                    & 28.4                    & 27.8                    & 37.2                     \\ 

                      & \textbf{ROME}     & 40.2                    & 26.9                    & 41.1                    & 25.2                    & 36.7                    & 23.5                    & 42.2                    & 26.6                    & 29.0                    & 37.5                     \\ 

                      & \textbf{KN}       & 28.3                    & 16.5                    & 29.6                    & 13.6                    & 27.2                    & 15.2                    & 31.0                    & 17.0                    & 18.2                    & 26.5                     \\
\hline
    \end{tabular}}
    \caption{Experimental Results of CounterFact with Gradient-based Baseline Methods. F1 scores are reported.}
    \tablabel{baseline_results}
\end{table*}

\section{Further Discussion on Related Work}
\applabel{further_discussion}
As discussed in \secref{related_work}, a few previous studies have explored gradient-free knowledge editing methods, notably ReMaKE~\citep{wang2023retrieval}. Our work diverges from these existing approaches in several key aspects, which we will elaborate on below. 
Regarding task setup, ReMaKE employs a cross-lingual retrieval-augmented strategy tailored for batch edits, allowing simultaneous modifications of multiple knowledge pieces, such as an entire knowledge base. Conversely, our approach focuses on individual knowledge edits, which do not involve cross-lingual retrieval. This fundamental difference sets our approach apart and addresses a unique aspect of knowledge editing.
In prompt engineering, when designing the demonstrations of ICL, ReMaKE uses translation pairs of source and target language facts to make up the cross-lingual demonstrations, from which the model cannot learn real cross-lingual knowledge editing competencies like portability and locality. In contrast, our MIKE method provides four different types of cross-lingual ICL demonstrations. From these demonstrations, the model can effectively learn real cross-lingual knowledge editing.

\section{Experiment Implementation Details}
\applabel{setting}

\tabref{setting} displays the experiment implementation details. We use the models from HuggingFace (\url{https://huggingface.co/meta-llama/Llama-3.2-3B}, \url{https://huggingface.co/meta-llama/Llama-3.1-8B}).

\section{Additional Experiments with Different Models}
To investigate the generalization of BMIKE-53, we also benchmark several more multilingual and English-centric LLMs, including Bloomz-7b1~\footnote{\url{https://huggingface.co/bigscience/bloomz-7b1}}, Ministral-8B~\footnote{\url{https://huggingface.co/mistralai/Ministral-8B-Instruct-2410}}, and Qwen2.5-7B~\footnote{\url{https://huggingface.co/Qwen/Qwen2.5-7B-Instruct}}.
We use a subset of 10 languages in BMIKE-53, and report the average results across the following 10 languages: de, es, fa, fr, hu, ja, ko, nl, ru, tr, vi, zh.
\tabref{add_exp} shows the results of the additional experiments. For better comparison, we also report the average results across the 10 languages for Llama3.1-8B and Llama3.2-3B.

\begin{table*}
\centering
\scalebox{0.76}{
\begin{tabular}{llcccccccccccc} 
\toprule
   &  & & \multicolumn{2}{c}{\textbf{zsre}}  & & &\multicolumn{2}{c}{\textbf{CounterFact}}     & &  &       \multicolumn{2}{c}{\textbf{WikiFactDiff}}   &  \\ 
\cline{3-14}
&     & \textbf{rel} &\textbf{gen} & \textbf{loc} & \textbf{port} & \textbf{rel} & \textbf{gen} & \textbf{loc} & \textbf{port} & \textbf{rel} & \textbf{gen} & \textbf{loc} & \textbf{port}  \\ 
\midrule
\multirow{4}{*}{\textbf{Bloomz-7b1}}   & zero-shot                & 27.13                                           & 26.53                                           & 1.85                                            & 4.40                                             & 24.78                                           & 23.97                                           & 5.53                                            & 6.62                                             & 28.68                                           & 28.42                                           & 3.00                                            & 1.56                                              \\ 
& one-shot                 & 28.71                                           & 27.94                                           & 1.06                                            & 3.57                                             & 26.21                                           & 25.40                                           & 4.84                                            & 5.59                                             & 24.54                                           & 24.28                                           & 2.10                                            & 1.10                                              \\ 
& 8-shot (mix)             & 31.26                                           & 30.86                                           & 2.42                                            & 5.87                                             & 31.86                                           & 31.33                                           & 8.29                                            & 6.97                                             & 26.87                                           & 26.78                                           & 4.13                                            & 2.48                                              \\ 
& 8-shot (metric)          & 33.83                                           & 33.43                                           & 2.35                                            & 6.20                                             & 34.85                                           & 34.43                                           & 11.94                                           & 8.93                                             & 32.50                                           & 32.45                                           & 4.62                                            & 3.46                                              \\ 
\midrule
\multirow{4}{*}{\textbf{Ministral-7b}} & zero-shot                & 50.63                                           & 49.30                                           & 8.62                                            & 12.16                                            & 41.65                                           & 39.97                                           & 16.24                                           & 13.89                                            & 51.48                                           & 48.97                                           & 5.71                                            & 3.06                                              \\ 
 & one-shot                 & 73.21                                           & 72.76                                           & 12.74                                           & 21.40                                            & 77.69                                           & 76.27                                           & 14.68                                           & 26.22                                            & 53.54                                           & 52.71                                           & 4.17                                            & 2.86                                              \\ 
& 8-shot (mix)             & 73.49                                           & 73.06                                           & 14.97                                           & 28.62                                            & 68.55                                           & 66.54                                           & 20.77                                           & 30.67                                            & 54.71                                           & 53.80                                           & 5.84                                            & 7.31                                              \\ 
& 8-shot (metric)          & 73.48                                           & 73.44                                           & 15.38                                           & 31.77                                            & 75.61                                           & 74.12                                           & 42.31                                           & 43.88                                            & 59.58                                           & 58.89                                           & 9.31                                            & 11.78                                             \\ 
\midrule
\multirow{4}{*}{\textbf{Qwen2.5-7b}}   & zero-shot                & 41.76                                           & 41.32                                           & 2.92                                            & 8.82                                             & 40.23                                           & 38.12                                           & 3.98                                            & 11.28                                            & 43.08                                           & 39.07                                           & 3.36                                            & 2.62                                              \\ 
& one-shot                 & 60.27                                           & 59.15                                           & 1.57                                            & 11.51                                            & 65.28                                           & 62.06                                           & 8.54                                            & 16.35                                            & 54.89                                           & 50.67                                           & 3.76                                            & 2.44                                              \\ 
   & 8-shot (mix)             & 63.36                                           & 63.05                                           & 3.51                                            & 16.51                                            & 64.24                                           & 62.05                                           & 15.39                                           & 21.34                                            & 53.48                                           & 51.11                                           & 4.69                                            & 3.63                                              \\ 
 & 8-shot (metric)          & 63.71                                           & 63.03                                           & 9.73                                            & 20.01                                            & 66.30                                           & 64.24                                           & 35.24                                           & 29.98                                            & 58.52                                           & 57.58                                           & 7.65                                            & 5.83                                              \\ 
\midrule
\multirow{4}{*}{\textbf{Llama3.2-3b}}  & zero-shot                & 50.16                                           & 49.03                                           & 5.64                                            & 5.41                                             & 43.51                                           & 40.84                                           & 7.53                                            & 5.61                                             & 58.28                                           & 57.41                                           & 6.77                                            & 3.18                                              \\ 
 & one-shot                 & 71.57                                           & 71.25                                           & 7.78                                            & 14.97                                            & 68.34                                           & 68.02                                           & 6.58                                            & 16.30                                            & 66.32                                           & 65.93                                           & 6.31                                            & 3.06                                              \\ 
    & 8-shot (mix)             & 70.54                                           & 70.49                                           & 8.89                                            & 16.43                                            & 70.80                                           & 70.33                                           & 5.11                                            & 13.75                                            & 64.97                                           & 64.60                                           & 6.24                                            & 4.00                                              \\ 
   & 8-shot (metric)          & 70.94                                           & 70.91                                           & 12.23                                           & 22.97                                            & 67.57                                           & 67.14                                           & 31.61                                           & 31.20                                            & 67.77                                           & 67.48                                           & 9.14                                            & 10.72                                             \\ 
\midrule
\multirow{3}{*}{\textbf{Llama3.1-8b}}  & zero-shot                & 65.53                                           & 64.09                                           & 9.76                                            & 10.05                                            & 63.01                                           & 60.59                                           & 18.68                                           & 11.16                                            & 67.84                                           & 66.40                                           & 10.04                                           & 4.15                                              \\ 
& one-shot                 & 75.27                                           & 74.90                                           & 13.36                                           & 20.81                                            & 71.92                                           & 71.29                                           & 12.66                                           & 21.93                                            & 70.53                                           & 69.84                                           & 7.80                                            & 4.15                                              \\ 
& 8-shot (mix)             & 74.29                                           & 74.00                                           & 15.46                                           & 25.18                                            & 75.15                                           & 74.42                                           & 11.40                                           & 23.74                                            & 68.57                                           & 67.86                                           & 8.27                                            & 8.87                                              \\ 
& 8-shot (metric)          & 74.86                                           & 74.79                                           & 16.15                                           & 32.86                                            & 73.88                                           & 73.19                                           & 47.55                                           & 41.17                                            & 71.98                                           & 71.34                                           & 13.84                                           & 14.58  \\
\bottomrule
\end{tabular}}
\caption{Additional experimental results with more various models. Average results across 10 languages (de, es, fa, fr, hu, ja, ko, nl, ru, tr, vi, zh).}
\tablabel{add_exp}
\end{table*}

\section{Full Results}
We show the full experimental results for all three tasks in~\tabref{results_zsre} (zsRE), \tabref{results_counterfact} (CounterFact), and \tabref{results_wfd} (WFD), respectively.

\begin{table*}[t]
\centering
\scalebox{.9}{
\begin{tabular}{cllccccc} 
\toprule
\textbf{lid}   & \textbf{language}    & \textbf{Family}     & \multicolumn{1}{c}{\textbf{syn\_sim}} & \multicolumn{1}{c}{\textbf{pho\_sim}} & \multicolumn{1}{c}{\textbf{inv\_sim}} & \multicolumn{1}{c}{\textbf{gen\_sim}} & \multicolumn{1}{c}{\textbf{geo\_sim}}  \\ 
\midrule
af    & Afrikaans   & Indo-European (Germanic)   & 84.94         & 81.83         & 69.10         & 50.46         & 86.84          \\ 

ar    & Arabic      & Semitic    & 65.11         & 70.09         & 70.81         & 0.15          & 97.04          \\ 

az    & Azerbaijani & Turkic     & 52.00         & 81.83         & 67.86         & 0.19          & 96.96          \\ 

be    & Belarusian  & Indo-European (Slavic)     & 78.64         & 85.83         & 70.42         & 16.80         & 99.35          \\ 

bg    & Bulgarian   & Indo-European (Slavic)     & 85.78         & 85.83         & 68.38         & 13.73         & 99.01          \\ 

bn    & Bengali     & Indo-European (Indo-Aryan) & 58.36         & 76.30         & 74.38         & 12.71         & 88.96          \\ 

ca    & Catalan     & Indo-European (Romance)    & 87.30         & 85.83         & 75.22         & 10.64         & 99.66          \\ 

ceb   & Cebuano     & Austronesian               & 62.17         & 76.30         & 75.22         & 0.13          & 81.50          \\ 

cs    & Czech       & Indo-European (Slavic)     & 73.99         & 85.83         & 66.51         & 13.73         & 99.71          \\ 

cy    & Welsh       & Indo-European (Celtic)     & 71.90         & 81.83         & 77.85         & 13.73         & 99.99          \\ 

da    & Danish      & Indo-European (Germanic)   & 88.01         & 81.83         & 77.54         & 40.90         & 99.89          \\ 

de    & German      & Indo-European (Germanic)   & 90.26         & 80.60         & 76.28         & 54.49         & 99.76          \\ 

el    & Greek       & Indo-European (Hellenic)   & 78.31         & 95.35         & 64.76         & 15.03         & 98.96          \\ 

es    & Spanish     & Indo-European (Romance)    & 82.16         & 85.83         & 63.83         & 9.71          & 99.59          \\ 

et    & Estonian    & Uralic     & 77.35         & 85.83         & 66.94         & 0.23          & 99.45          \\ 

eu    & Basque      & Isolate    & 62.36         & 85.29         & 56.88         & 3.33          & 99.76          \\ 

fa    & Persian     & Indo-European (Iranian)    & 50.03         & 78.35         & 72.83         & 13.73         & 94.23          \\ 

fi    & Finnish     & Uralic     & 71.08         & 87.05         & 70.00         & 0.19          & 99.19          \\ 

fr    & French      & Indo-European (Romance)    & 81.18         & 75.28         & 74.09         & 9.71          & 99.93          \\ 

ga    & Irish       & Indo-European (Celtic)     & 72.01         & 85.83         & 69.35         & 12.71         & 99.96          \\ 

gl    & Galician    & Indo-European (Romance)    & 80.23         & 90.46         & 70.75         & 10.14         & 99.65          \\ 

he    & Hebrew      & Semitic    & 75.15         & 72.55         & 64.37         & 0.13          & 97.16          \\ 

hi    & Hindi       & Indo-European (Indo-Aryan) & 61.63         & 78.35         & 70.91         & 12.71         & 91.10          \\ 

hr    & Croatian    & Indo-European (Slavic)     & 83.18         & 85.83         & 69.67         & 12.71         & 99.50          \\ 

hu    & Hungarian   & Uralic     & 69.40         & 85.83         & 74.03         & 0.33          & 99.46          \\ 

hy    & Armenian    & Indo-European (Satem)      & 63.03         & 69.66         & 68.73         & 19.39         & 97.23          \\ 

id    & Indonesian  & Austronesian               & 72.66         & 90.92         & 75.58         & 0.12          & 79.16          \\ 

it    & Italian     & Indo-European (Romance)    & 85.78         & 85.83         & 70.00         & 11.21         & 99.53          \\ 

ja    & Japanese    & Isolate    & 50.03         & 66.77         & 65.40         & 0.19          & 85.65          \\ 

ka    & Georgian    & Caucasian  & 68.50         & 66.93         & 62.93         & 0.19          & 97.09          \\ 

ko    & Korean      & Isolate    & 55.29         & 74.65         & 70.94         & 0.33          & 86.93          \\ 

la    & Latin       & Indo-European (Romance)    & 78.27         & 85.83         & 76.76         & 15.03         & 99.47          \\ 

lt    & Lithuanian  & Indo-European (Baltic)     & 69.33         & 80.42         & 74.63         & 19.39         & 99.44          \\ 

lv    & Latvian     & Indo-European (Baltic)     & 75.39         & 81.83         & 75.22         & 19.39         & 99.42          \\ 

ms    & Malay       & Austronesian               & 70.49         & 90.92         & 72.49         & 0.15          & 80.49          \\ 

nl    & Dutch       & Indo-European (Germanic)   & 92.43         & 81.83         & 72.24         & 44.51         & 99.96          \\ 

pl    & Polish      & Indo-European (Slavic)     & 78.64         & 85.83         & 65.29         & 15.03         & 99.63          \\ 

pt    & Portuguese  & Indo-European (Romance)    & 84.24         & 90.46         & 78.68         & 10.14         & 99.68          \\ 

ro    & Romanian    & Indo-European (Romance)    & 79.60         & 90.46         & 73.42         & 11.89         & 99.22          \\ 

ru    & Russian     & Indo-European (Slavic)     & 81.18         & 85.83         & 64.76         & 16.80         & 95.81          \\ 

sk    & Slovak      & Indo-European (Slavic)     & 82.16         & 85.83         & 70.66         & 15.03         & 99.55          \\ 

sl    & Slovenian   & Indo-European (Slavic)     & 80.59         & 85.83         & 75.58         & 15.03         & 99.62          \\ 

sq    & Albanian    & Indo-European (Other)      & 79.60         & 87.05         & 72.49         & 33.48         & 99.19          \\ 

sr    & Serbian     & Indo-European (Slavic)     & 79.60         & 85.83         & 72.94         & 12.71         & 99.23          \\ 

sv    & Swedish     & Indo-European (Germanic)   & 93.34         & 81.83         & 67.98         & 40.90         & 99.62          \\ 

ta    & Tamil       & Dravidian  & 51.36         & 85.29         & 65.81         & 0.11          & 87.95          \\ 

th    & Thai        & Kra-Dai    & 63.95         & 78.35         & 74.91         & 0.11          & 85.25          \\ 

tr    & Turkish     & Turkic     & 50.68         & 81.83         & 66.59         & 0.14          & 98.25          \\ 

uk    & Ukrainian   & Indo-European (Slavic)     & 84.73         & 85.83         & 74.38         & 15.03         & 99.28          \\ 

ur    & Urdu        & Indo-European (Indo-Aryan) & 61.63         & 85.83         & 71.57         & 12.71         & 92.54          \\ 

vi    & Vietnamese  & Austroasiatic              & 66.04         & 78.35         & 74.74         & 0.19          & 85.25          \\ 

zh-cn & Chinese     & Sino-Tibetan               & 71.08         & 72.55         & 69.73         & 0.33          & 88.42          \\
\bottomrule
\end{tabular}}
\caption{Detailed information of the languages covered by BMIKE-53. The right five columns show the linguistic feature similarities between the target language and English. syn: syntax, pho: phonology, inv: phonetics, gen: phylogenetic, geo: geographic, sim: similarity.}
\tablabel{lang_info}
\end{table*}

\begin{figure*}[t]
    \centering
    \includegraphics[height=.95\textheight]{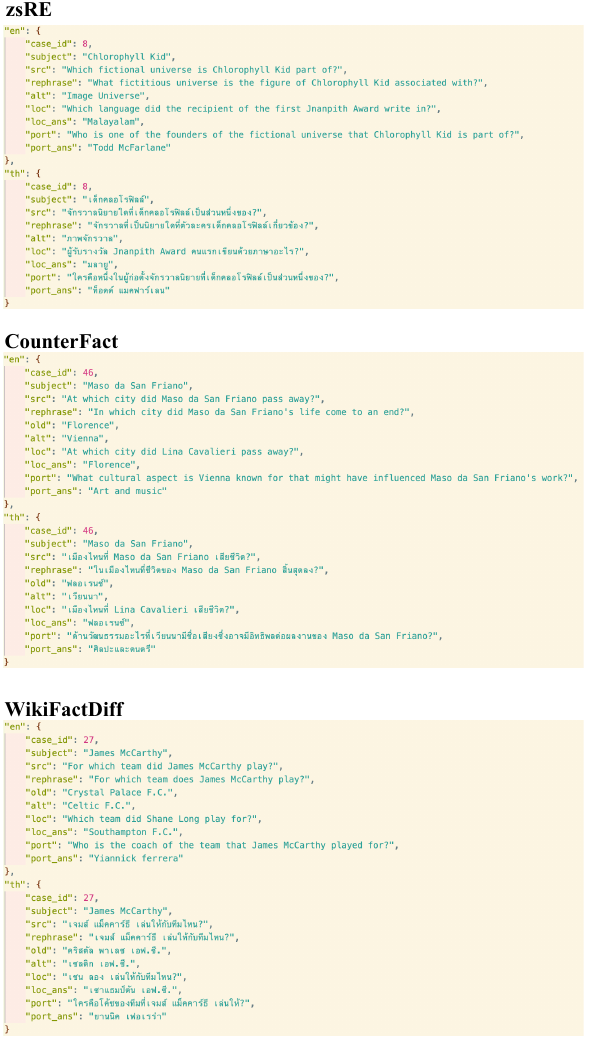}
    \caption{Data Item Examples of BMIKE-53.}
    \figlabel{example}
\end{figure*}

\begin{table*}
    \centering
    \includegraphics[width=\textwidth]{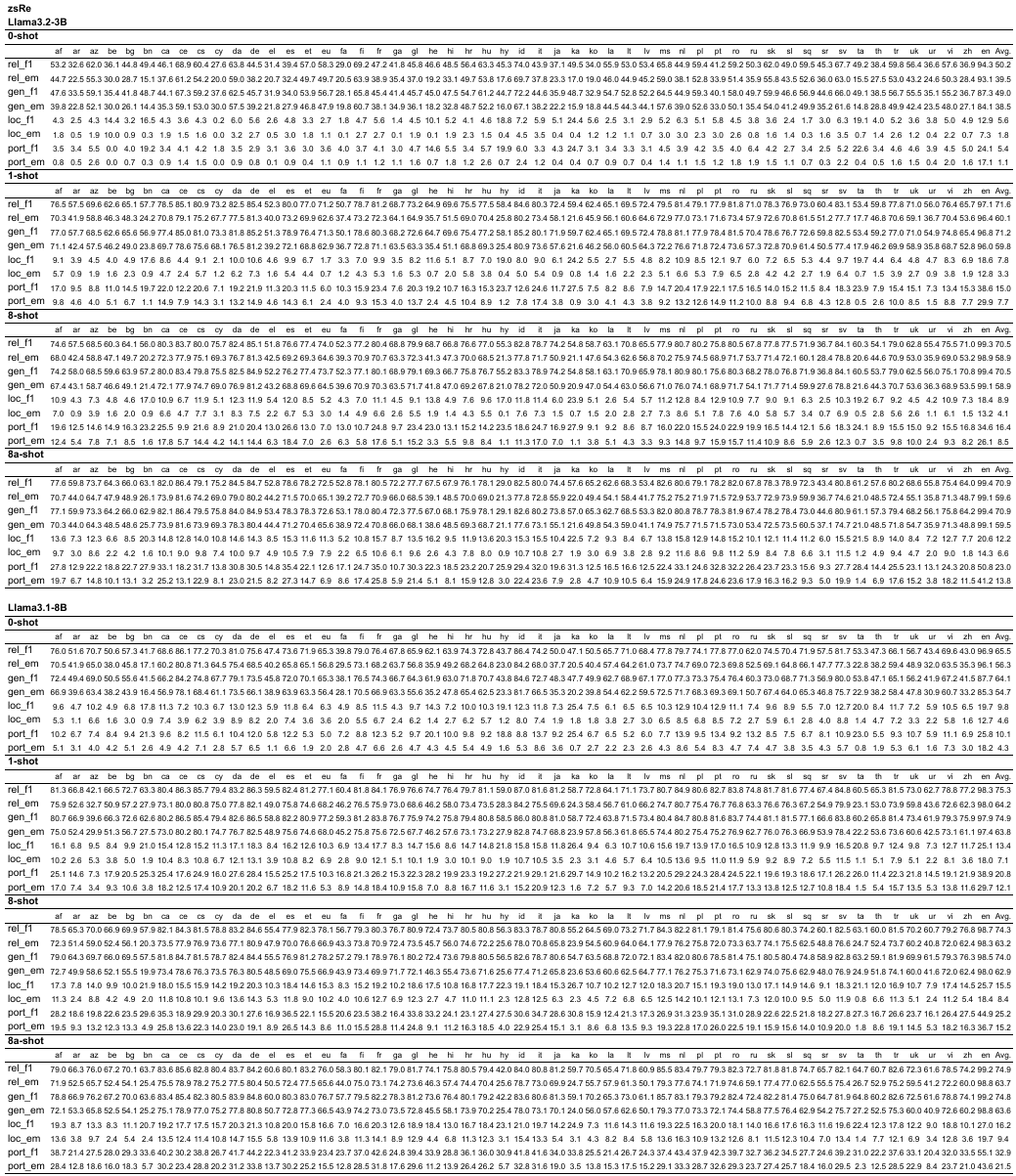}
    \caption{Full Results on zsRE.}
    \tablabel{results_zsre}
\end{table*}

\begin{table*}
    \centering
    \includegraphics[width=\textwidth]{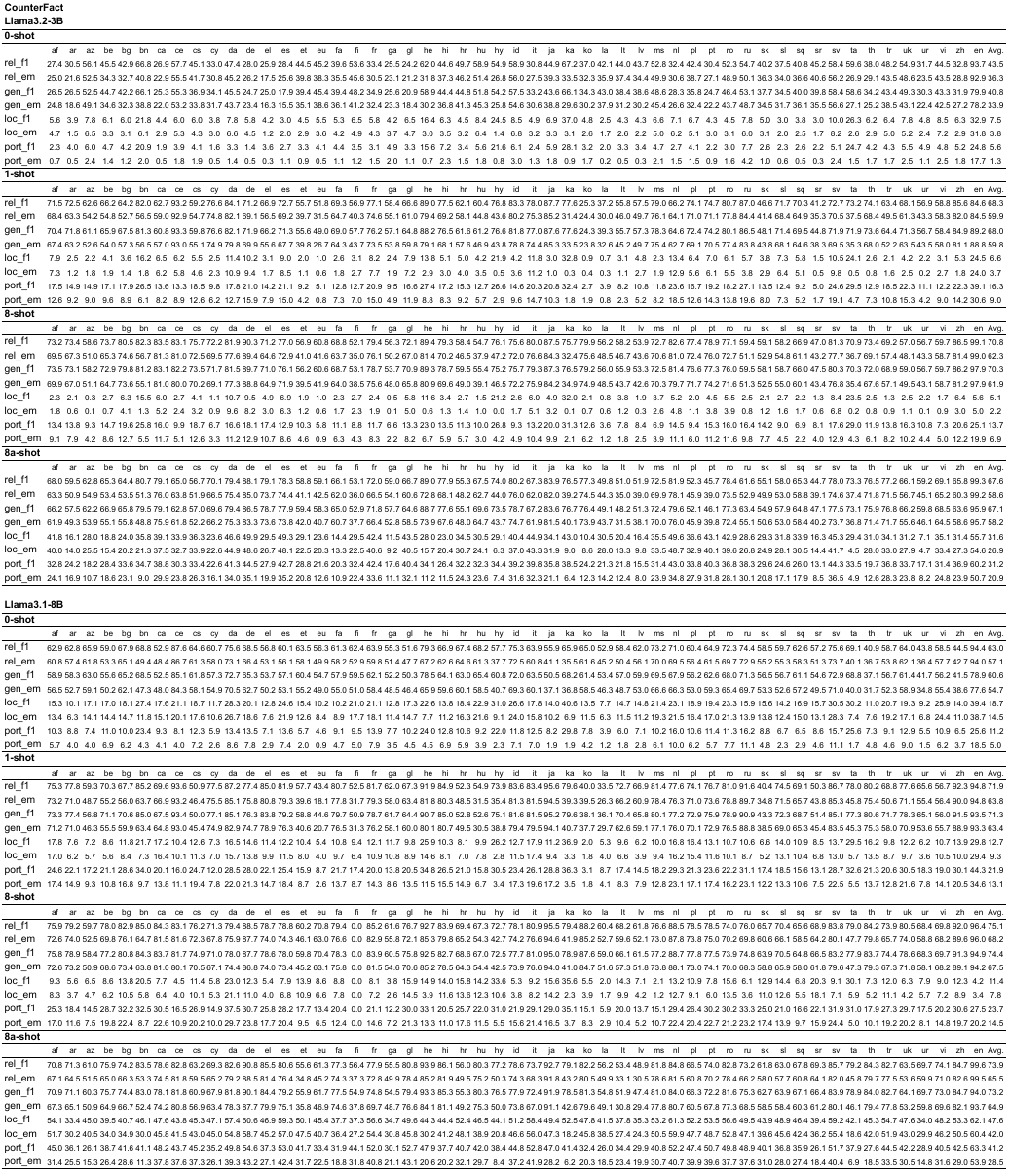}
    \caption{Full Results on CounterFact.}
    \tablabel{results_counterfact}
\end{table*}

\begin{table*}
    \centering
    \includegraphics[width=\textwidth]{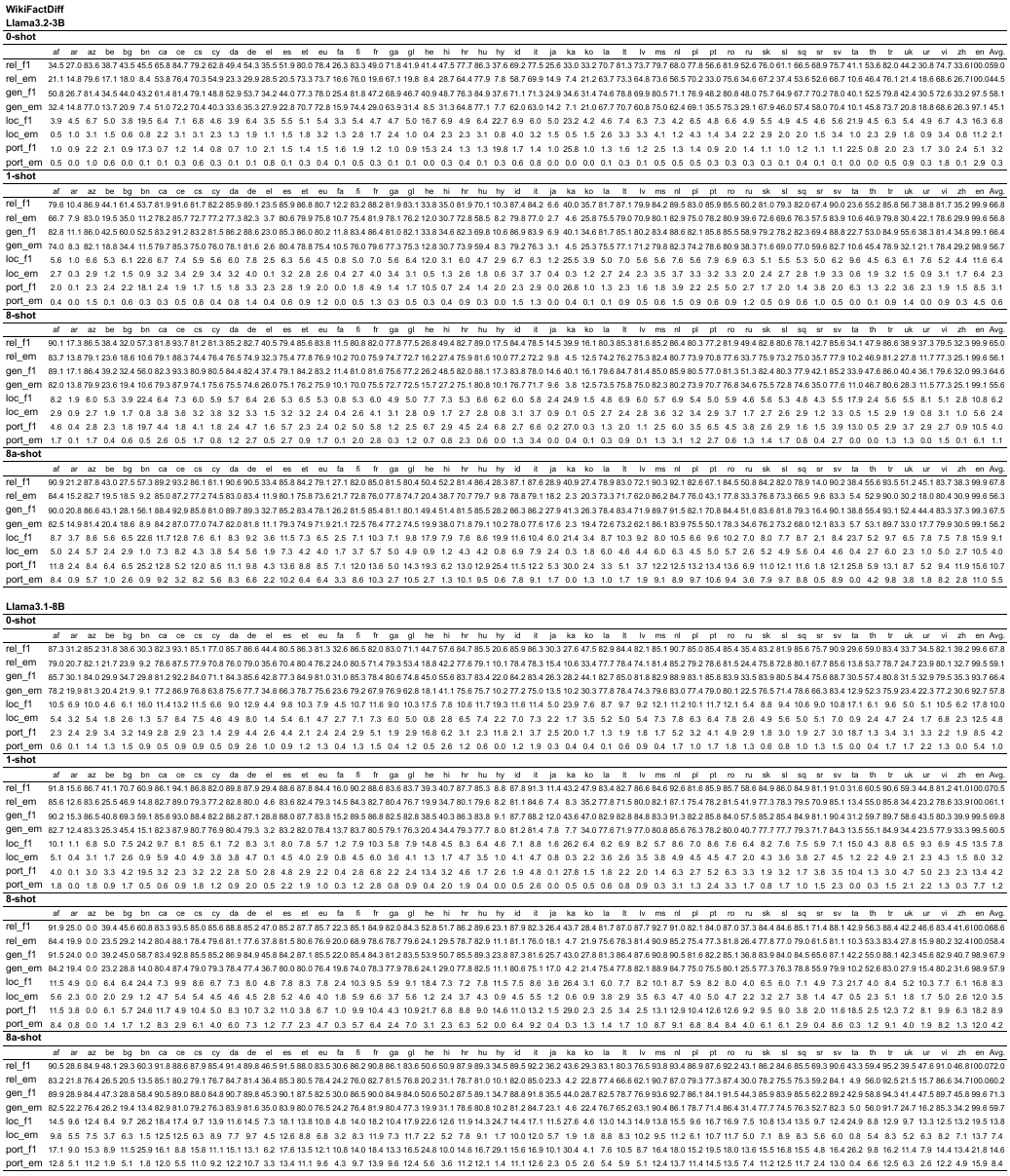}
    \caption{Full Results on WFD.}
    \tablabel{results_wfd}
\end{table*}

\end{document}